\documentclass{article}

\usepackage{microtype}
\usepackage{graphicx}
\usepackage{booktabs}
\usepackage{array}
\usepackage{amsmath}
\usepackage{amssymb}
\usepackage{mathtools}
\usepackage{xurl}
\usepackage{hyperref}
\usepackage[capitalize,noabbrev]{cleveref}

\usepackage[accepted]{icml2026}

\makeatletter
\renewcommand{\ICML@appearing}{\textit{Accepted at the ICML 2026 Workshop on
Hypothesis Testing}, 2026.}
\makeatother

\icmltitlerunning{Metadata Predictability Is Not Evidence Dependence}

\begin{document}

\twocolumn[
\icmltitle{Metadata Predictability Is Not Evidence Dependence:
An Intervention-Based Audit for Weak-Label Benchmarks}

\begin{icmlauthorlist}
\icmlauthor{Kan Shao}{jinglue}
\end{icmlauthorlist}

\icmlaffiliation{jinglue}{Jinglue Technology Development (Nanjing) Co., Ltd., Nanjing, China}
\icmlcorrespondingauthor{Kan Shao}{shaokan1991@gmail.com}

\icmlkeywords{benchmark audit, weak supervision, hypothesis testing, evaluation, shortcut learning}

\vskip 0.3in
]

\printAffiliationsAndNotice{}

\begin{abstract}
We study a protocol-level test for weak-label benchmarks: whether benchmark
outputs change when the provided evidence is intervened on. Metadata-only
shortcut checks answer a different question, namely whether outputs are
predictable from metadata priors. We therefore combine a metadata statistic,
the Metadata Prior Dominance Score (MPDS), with an evidence-intervention
statistic, $\Delta$Evi, measuring sensitivity to evidence identity under
cross-item shuffling. Synthetic HotpotQA gives a
constructed counterexample to metadata-only screening: MPDS is only
moderate (0.643), yet
$\Delta$Evi is zero. Stronger-reader reruns show why calibration belongs in the
test procedure: SNLI shows a calibration reversal, reconstructed HotpotQA
occupies a question-dominant warning region, and FEVER is a strongly
evidence-sensitive positive control across four transformers. The practical
lesson is simple: benchmark audits should
report metadata-only screening, evidence intervention, and reader-strength
calibration together.
\end{abstract}

\section{Introduction}

Consider a weak-label protocol that generates benchmark labels from heuristics
over query types, answer forms, or claim patterns. The usual audit question is
whether those labels are predictable from metadata alone. For evidence-based
evaluation, the sharper question is:
\emph{does the protocol actually depend on the provided evidence?} This concern
fits broader critiques of benchmark validity and evaluation practice in
NLP~\cite{ethayarajh2020utility,bowman2021fixbenchmarking,kiela2021dynabench}.
It also matches recent work that treats robustness and factuality evaluation as
statistical testing problems~\cite{rauba2025statistical,nie2025facttest} and
develops testing tools such as e-values for prediction-assisted
inference~\cite{csillag2025predictionpowered}.

We frame this mismatch as a hypothesis-testing problem. Metadata-only
predictability is informative, but it does not test the null hypothesis that
protocol behavior is invariant to evidence identity. We therefore combine
MPDS, a metadata-prior accuracy ratio, with the evidence-shuffling statistic
$\Delta$Evi, which measures how much performance changes when evidence
identity is broken. The audit therefore returns two decision statistics: MPDS
for metadata predictability and $\Delta$Evi for evidence dependence.

This framing adds a protocol-level layer to familiar dataset-artifact and
model-shortcut analyses. Dataset-artifact work asks whether data collection
leaves shortcut cues~\cite{gururangan2018annotation}. Model-shortcut analyses
ask whether systems exploit heuristic cues~\cite{mccoy2019right,ribeiro2020checklist}. We
ask a third question: whether the \emph{evaluation protocol itself} rewards
metadata recovery rather than evidence use. This connects weak-label benchmark
construction to data programming, weak-supervision systems, and weak-supervision
benchmark suites~\cite{ratner2016data,ratner2018snorkel,zhang2021wrench,zhang2024stronger},
but focuses on evaluation integrity rather than only label efficiency.
It also complements reproducibility-oriented benchmark audits and
partial-identification approaches to weak-supervision
evaluation~\cite{calamai2025benchmarking,polo2024partialidentification}.

The empirical result is a compact diagnostic map. Synthetic HotpotQA yields
a constructed latent-coupling counterexample: MPDS is
moderate, yet $\Delta$Evi is zero. Real-benchmark audits show why
stronger-reader calibration belongs in the test procedure: SNLI shows a
calibration reversal, and reconstructed HotpotQA highlights a
warning region. FEVER provides a positive control:
its labels are clearly evidence-sensitive under calibrated readers. OOD and
counterfactual analyses show that unstable protocol behavior can also have
downstream consequences.

\paragraph{Contributions.}
We make three contributions: we separate metadata predictability from evidence
dependence, define a two-statistic audit using MPDS and $\Delta$Evi, and show a
compact diagnostic map with four illustrative cases: constructed
counterexample, calibration reversal, positive control, and warning case.

\section{A Two-Statistic Test for Protocol Dependence}

The test has two axes. Let $\mathrm{Acc}_{\rm meta}$ be the accuracy of a
metadata-majority predictor and $\mathrm{Acc}_{\rm full}$ be the accuracy of
the full audited system. MPDS normalizes metadata predictability by full-system
accuracy. For evidence dependence, the null hypothesis is that behavior is
invariant to evidence identity:
\[
\begin{aligned}
H_0 &: \mathrm{Acc}_{\rm full} = \mathrm{Acc}_{\rm shuf},\\
\mathrm{MPDS} &:= \mathrm{Acc}_{\rm meta}/\mathrm{Acc}_{\rm full},\\
\Delta\mathrm{Evi} &:=
\mathrm{Acc}_{\rm full}-\mathrm{Acc}_{\rm shuf},
\end{aligned}
\]
where $\mathrm{Acc}_{\rm shuf}$ uses a cross-item evidence permutation while
queries and labels are fixed. This is a paired intervention on evidence
identity: the question and target label remain the same, but the evidence
attached to the item is replaced. Near-zero $\Delta$Evi indicates invariance to
evidence identity; positive $\Delta$Evi indicates evidence-sensitive behavior.
In practice we estimate $\mathrm{Acc}_{\rm shuf}$ over $K$ independent
evidence permutations and report the mean and per-permutation population standard
deviation $\sigma_{\rm shuf}$ (population SD, not standard error); it is small
because the intervention collapses predictions to a near-constant distribution
across permutations (under shuffled evidence the $K=8$ FEVER model abstains on
1980--1994 of 2000 items in \emph{every} permutation), not by construction. An
independent $K=50$ BERT rerun gives $\sigma_{\rm shuf}\approx0.004$ for both FEVER
and SNLI---roughly $3\times$ the $\approx0.0015$ that a standard error of the mean
would give, and below the i.i.d.\ single-shuffle bound ($\approx0.010$)---so the
reported deviations are genuine per-permutation SDs, not mislabeled SEMs. We use
$K=8$ for the reported sweep and recommend $K\ge 20$ for production audits.

Together, these statistics define a diagnostic map with three illustrative
regions:
\textbf{direct coupling}
(high MPDS, near-zero $\Delta$Evi), \textbf{latent coupling} (moderate MPDS,
near-zero $\Delta$Evi), and \textbf{evidence-sensitive protocols} (clearly
positive $\Delta$Evi). These are not exhaustive categories but useful signposts.
The latent-coupling region is the critical one: the
metadata screen is not dominant enough to look trivial, but the evidence
intervention still shows no dependence on evidence identity.
MPDS as a ratio conflates metadata strength with task difficulty
(e.g., $(0.5,0.5)$ and $(0.8,0.8)$ both give $1.0$). A chance-corrected form
$\mathrm{MPDS}_{+}=(\mathrm{Acc}_{\rm meta}-c)/(\mathrm{Acc}_{\rm full}-c)$
($c$ the majority rate) need not track it: synthetic HotpotQA has ratio $0.643$
but near-zero lift over chance ($\mathrm{MPDS}_{+}=0.03$), while NQ is $1.0$ under
both, and $\mathrm{MPDS}_{+}$ is itself ill-conditioned when
$\mathrm{Acc}_{\rm full}\!\approx\!c$. We report the ratio as the screen and read
both only alongside $\Delta$Evi. For real
benchmarks, we treat lightweight TF-IDF+LR as
a \emph{screening layer}, stronger transformer reruns as a \emph{calibration
layer}, and OOD or counterfactual analyses as \emph{consequence evidence}.
The decision rule is deliberately operational: a near-zero
$\Delta$Evi at the screening layer triggers calibration; persistent near-zero
$\Delta$Evi after calibration is a warning region; consistently positive
$\Delta$Evi rejects evidence invariance for the audited reader family.

We instantiate the audit on controlled synthetic HotpotQA (a constructed
counterexample built to exhibit latent coupling) and three
evidence-bearing benchmark settings: SNLI~\citep{bowman2015snli},
FEVER~\citep{thorne2018fever}, and reconstructed HotpotQA~\citep{yang2018hotpotqa}.
Reconstructed HotpotQA uses the HuggingFace \texttt{fullwiki} config
(train $=$ 2000, eval $=$ 600) with all retrieved Wikipedia paragraphs as
evidence; labels use a heuristic over question type, answer type, and
supporting-fact count.
Calibration uses four transformer families where shown, and input ablations
separate evidence sensitivity from residual query- or hypothesis-side signal.
For a new weak-label benchmark, the audit is a four-step packet: specify the
metadata schema used by the protocol, compute MPDS as a metadata-only screen,
estimate $\Delta$Evi under paired evidence shuffles, and rerun near-zero cases
with stronger readers plus input ablations.

\section{Results}

\subsection{Controlled and Real-Benchmark Test Outcomes}

The constructed counterexample is synthetic HotpotQA, where MPDS is only 0.643 but
$\Delta$Evi is zero. This is the key latent-coupling case:
metadata-only screening would look moderately reassuring, yet evidence identity
has no measured effect.
The synthetic suite also supplies endpoints for the decision map: a synthetic
NQ-style task is a direct-coupling stress test (MPDS $=1.0$, $\Delta$Evi $=0$),
while a synthetic TriviaQA-style task is evidence-sensitive
($\Delta$Evi $=0.808$).

The lightweight real-benchmark layer is mixed. SNLI and reconstructed real
HotpotQA both show near-zero $\Delta$Evi under TF-IDF+LR, but for different
reasons: a weak-reader limitation in SNLI, and question-dominant collapse under
severe skew in reconstructed HotpotQA. FEVER, by contrast, is already positive under LR.
The lightweight layer is therefore a screening stage rather than the final
decision.

\begin{table}[t]
\centering
\scriptsize
\setlength{\tabcolsep}{3pt}
\begin{tabular}{>{\raggedright\arraybackslash}p{0.20\linewidth}>{\raggedright\arraybackslash}p{0.34\linewidth}>{\raggedright\arraybackslash}p{0.36\linewidth}}
\toprule
Case & Test outcome & Why it matters \\
\midrule
HotpotQA (syn.) & MPDS $=0.643$, $\Delta$Evi $=0$ & metadata screen can miss evidence independence \\
SNLI & LR $\approx 0$; transformers $=0.26$--$0.37$ & weak-reader false negative; calibration changes the conclusion \\
FEVER & LR $\Delta$Evi $=0.13$; transformers $=0.63$--$0.68$ & evidence-sensitive positive control \\
HotpotQA (recon.) & q-only $=0.975$; BERT-like $\approx 0$ & query dominance, skew, and reader collapse \\
\bottomrule
\end{tabular}
\caption{Decision view of the audit outcomes. MPDS and $\Delta$Evi separate
metadata predictability, evidence sensitivity, and reader calibration effects.}
\label{tab:icml_summary}
\end{table}

These four cases span the diagnostic spectrum, from the latent-coupling
counterexample through calibration reversal and the question-dominant warning
region to the positive control.

\subsection{Calibration Changes the Test Conclusion}

SNLI is the clearest calibration example. Under the lightweight reader, $\Delta$Evi is
near zero ($-0.010 \pm 0.005$ over $K=8$ premise-level shuffles). Under stronger
readers, that conclusion is overturned consistently:
multishuffle BERT yields $\Delta$Evi$=0.3671 \pm 0.0036$,
DistilBERT $0.2954 \pm 0.0051$, ELECTRA-small $0.2558 \pm 0.0039$, and
SciBERT $0.2639 \pm 0.0061$. Under calibrated readers, SNLI is therefore not
evidence-independent. At the same time, SciBERT input ablations show that
hypothesis-only signal remains strong (accuracy $0.5975$; premise-only
$0.3365$), so the right conclusion is
\emph{evidence-sensitive after calibration, with residual hypothesis-side signal}.

FEVER \citep{thorne2018fever} is the positive-control case. Lightweight LR
already has $\Delta$Evi$=0.13$, and the transformer sweep is strongly positive:
BERT
$0.6813 \pm 0.0022$, DistilBERT $0.6423 \pm 0.0028$,
ELECTRA-small $0.6428 \pm 0.0042$, and SciBERT $0.6580 \pm 0.0025$
with a second SciBERT seed at $0.6338 \pm 0.0038$. Thus the audit identifies
both evidence-insensitive and strongly evidence-sensitive protocols.

Reconstructed HotpotQA remains near-zero across BERT, DistilBERT, and
ELECTRA-small ($\Delta$Evi $\le 0.002$, $\sigma_{\rm shuf} \le 0.002$).
The label distribution is severely skewed (578 \textsc{full} vs.\
22 \textsc{conflict} in eval), and the question-only baseline reaches
$0.975$, so near-zero $\Delta$Evi reflects question-side collapse
rather than clean evidence independence. It remains a warning case.

\begin{figure}[t]
\centering
\includegraphics[width=0.98\linewidth]{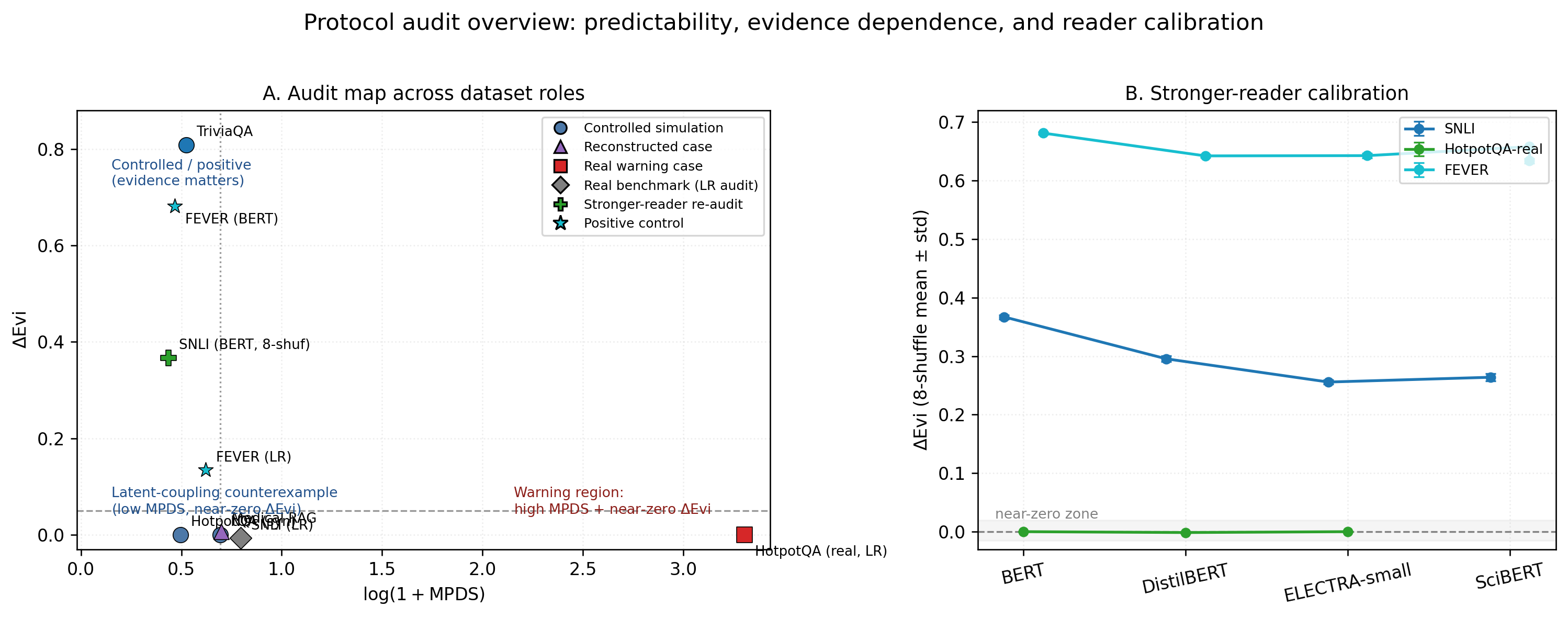}
\caption{Diagnostic map under the intervention-based audit view. Left: MPDS and
$\Delta$Evi separate controlled positive cases, the synthetic HotpotQA
latent-coupling counterexample, the FEVER positive control, and the
reconstructed HotpotQA warning region. Right: stronger-reader reruns show that SNLI and
FEVER are positive after calibration, whereas reconstructed real
HotpotQA stays in the warning region.}
\label{fig:icml_map}
\end{figure}

\subsection{Consequences Under Distribution Shift}

Protocol coupling has measurable downstream consequences. Under OOD
answer-type shift, the synthetic NQ-style task collapses completely; SNLI and
both HotpotQA variants show degradation tied to metadata dimensions driving
protocol behavior. Counterfactual metadata flips are strongest for the
synthetic NQ task (flipping answer-type changes every held-out label with
evidence fixed), weaker for HotpotQA due to \textsc{full}-label dominance.
MPDS-gated filtering is not a reliable fix: on synthetic HotpotQA, removing
the dominant high-risk group worsens the OOD gap, showing post-hoc deletion
is insufficient once the shortcut is built into the protocol.

\section{Discussion}

The main lesson is methodological: metadata predictability is not evidence
dependence, so an audit should report a metadata screen, an evidence-intervention
statistic, and a calibrated stronger-reader rerun together---stating the metadata
schema, shuffle count, and reader family---rather than a single shortcut baseline.

\section{Limitations}

Our sweep is budget-limited (four transformers, $K=8$ permutations; $K\ge 20$
is preferable). Metadata features are hand-designed, so higher-order couplings
may escape detection. MPDS as a simple ratio conflates metadata strength with
task difficulty. The three-region diagnostic map is illustrative, not exhaustive.
The synthetic HotpotQA counterexample is constructed, and we found no natural
benchmark in the latent-coupling region; reconstructed HotpotQA is shaped by
severe label skew (96\% majority). The framework targets evidence-identity
sensitivity rather than semantic reasoning quality.

\bibliography{refer_papers}
\bibliographystyle{icml2026}

\end{document}